\documentclass[letterpaper]{article} 
\usepackage{aaai25}  
\usepackage{times}  
\usepackage{helvet}  
\usepackage{courier}  
\usepackage[hyphens]{url}  
\usepackage{graphicx} 
\usepackage{natbib}  
\usepackage{caption} 
\usepackage{algorithm}
\usepackage{algorithmic}

\usepackage{newfloat}
\usepackage{listings}
\DeclareCaptionStyle{ruled}{labelfont=normalfont,labelsep=colon,strut=off} 
\floatstyle{ruled}
\newfloat{listing}{tb}{lst}{}
\floatname{listing}{Listing}
\usepackage[utf8]{inputenc} 
\usepackage[T1]{fontenc}    
\usepackage{booktabs}       
\usepackage{amsfonts}       
\usepackage{nicefrac}       
\usepackage{microtype}
\usepackage{subcaption}
\usepackage{amsmath}
\usepackage[title]{appendix}

\usepackage{colortbl}
\usepackage{xcolor}

\usepackage{appendix}

\title{ChemVLM: Exploring the Power of Multimodal Large Language Models in Chemistry Area}

\author{
    Junxian Li\textsuperscript{\rm 1, 2}
    Di Zhang\textsuperscript{\rm 2, 3}
    Xunzhi Wang\textsuperscript{\rm 2, 4}
    Zeying Hao\textsuperscript{\rm 2, 5}
    Jingdi Lei\textsuperscript{\rm 2},
    Qian Tan\textsuperscript{\rm 2, 5},
    Cai Zhou\textsuperscript{\rm 2},
    Wei Liu\textsuperscript{\rm 1, 2},
    Yaotian Yang\textsuperscript{\rm 2},
    Xinrui Xiong\textsuperscript{\rm 2},
    Weiyun Wang\textsuperscript{\rm 2, 3}, Zhe Chen\textsuperscript{\rm 2}, Wenhai Wang\textsuperscript{\rm 2}, Wei Li\textsuperscript{\rm 2}, Mao Su\textsuperscript{\rm 2}, Shufei Zhang\textsuperscript{\rm 2}, Wanli Ouyang\textsuperscript{\rm 2}, Yuqiang Li\textsuperscript{\rm 2}\footnotemark[2], Dongzhan Zhou\textsuperscript{\rm 1}\thanks{Corresponding author} 
}

\affiliations{
    \textsuperscript{\rm 1}Shanghai Jiao Tong University, \\
    \textsuperscript{\rm 2}Shanghai Artificial Intelligence Laboratory, \\
    \textsuperscript{\rm 3}Fudan University, \\
    \textsuperscript{\rm 4}Nankai University, \\
    \textsuperscript{\rm 5}University of Science and Technology of China, \\

    lijunxian0531@sjtu.edu.cn, di.zhang@ustc.edu, xunzhi@mail.nankai.edu.cn, haozeying@mail.ustc.edu.cn, \\
    \{liyuqiang, zhoudongzhan\}@pjlab.org.cn

}

\usepackage{bibentry}

\begin{document}
\maketitle


\begin{abstract}

Large Language Models (LLMs) have achieved remarkable success and have been applied across various scientific fields, including chemistry. However, many chemical tasks require the processing of visual information, which cannot be successfully handled by existing chemical LLMs. This brings a growing need for models capable of integrating multimodal information in the chemical domain. In this paper, we introduce \textbf{ChemVLM}, an open-source chemical multimodal large language model specifically designed for chemical applications. ChemVLM is trained on a carefully curated bilingual multimodal dataset that enhances its ability to understand both textual and visual chemical information, including molecular structures, reactions, and chemistry examination questions. We develop three datasets for comprehensive evaluation, tailored to Chemical Optical Character Recognition (OCR), Multimodal Chemical Reasoning (MMCR), and Multimodal Molecule Understanding tasks. We benchmark ChemVLM against a range of open-source and proprietary multimodal large language models on various tasks. Experimental results demonstrate that ChemVLM achieves competitive performance across all evaluated tasks. 

\end{abstract}

\begin{links}
\link{Code}{https://github.com/lijunxian111/ChemVlm}
\link{Training Data}{https://huggingface.co/datasets/di-zhang-fdu/chemvlm-sft-datasets}
\link{Test Data}{https://github.com/lijunxian111/ChemVlm}
\end{links}

%

\section{Introduction}

Large Language Models (LLMs) have been widely adopted in various scientific domains due to their high potential to accelerate scientific discovery~\cite{zhang2024chemllm, m2024augmenting, jablonka2024leveraging,chemdfm, dagdelen2024structured}. However, despite their promising potential to advance research, purely text-based models are limited in handling the diverse multimodal data encountered in the fields of chemistry, including molecular structures, reaction equations, and other related phenomena. These limitations can impede the performance of comprehensive multimodal chemical reasoning (MMCR) tasks, leading to potential misinterpretations or misleading hallucinations. Moreover, chemists often spend significant time manually redrawing chemical images using software like ChemDraw~\cite{brown2014chemdraw} to convert them into machine-readable formats such as SMILES~\cite{weininger1988smiles} or IUPAC names~\cite{10.1039/9781849733069}.  Although traditional Chemical Optical Character Recognition (OCR) models, especially transformer-based ones like MolScribe~\cite{ImageRecog1} and Decimer~\cite{ImageRecog3}, have achieved remarkable success in converting chemical images into SMILES~\cite{ImageRecog1, rajan2021decimer}, their capabilities are limited to modality conversion and thus fails to address multimodal chemical reasoning (MMCR) tasks. For MMCR tasks, naive modality conversion was inadequate indeed, a more sophisticated capability that can offer a deeper insight into the non-text data is still required for the community. Therefore, a model that can seamlessly bridge visual and textual information in chemistry is desired.

While existing multimodal Large Language Models (MLLMs)~\cite{llava, llavanext, gpt4v, qwenvl, young2024yi, wang2023cogvlm} excel at processing diverse data types, their generic nature falls short in the chemical domain due to limited specialized knowledge. This gap necessitates a model specifically designed for chemistry that integrates textual and visual information, thereby enhancing research and analytical efficiency for chemists, overcoming the limitations of generalist approaches by providing precise, domain-specific insights.


To address this issue, we introduce ChemVLM\footnote{This work was done during their internship at Shanghai Artificial Intelligence Laboratory. }, an open-source multimodal large language model specifically designed for the chemical domain, built upon the ViT-MLP-LLM architecture~\cite{llava}. Our model combines a state-of-the-art Vision Transformer (ViT)~\cite{dosovitskiy2020image} for robust image encoding with ChemLLM~\cite{zhang2024chemllm}, a domain-specific Large Language Model trained on billions of tokens from high-quality chemical data. This architecture effectively bridges the gap in domain-specific knowledge needed to process multimodal chemical tasks. A Multi-Layer Perceptron (MLP) layer is adopted as a projector to perform image-text alignment. We implement a two-stage supervised fine-tuning strategy to enhance model performance. In the first stage, we freeze the LLM component to focus on modality alignment. In the second stage, we unfreeze and update all parameters to adapt the model to downstream tasks. This structured approach ensures that ChemVLM achieves competitive performance across a variety of chemical tasks.

To rigorously evaluate the performance of chemical domain models, we introduce three specialized datasets: ChemOCR, MMCR-Bench, and MMChemBench. Each dataset is designed to assess distinct aspects of model capabilities. Specifically, ChemOCR focuses on modality conversion, offering bilingual questions that challenge models to recognize the SMILES format from molecular images, directly testing OCR capabilities within a chemical context. MMCR-Bench is mainly derived from the chemistry section of the Chinese college entrance examination and expands to assess models on their ability to solve complex multimodal chemical reasoning (MMCR) problems. MMChemBench, extended from ChemBench~\cite{zhang2024chemllm}, includes molecule caption and molecule property prediction tasks with multimodal information, providing a comprehensive view of a model's capability on Multimodal Molecule Understanding. Additionally, we further validate our model’s performance using established open-source benchmarks~\cite{lu2022learn, he2024cmmu}, ensuring a comprehensive assessment across diverse chemical tasks.

In our assessment, ChemVLM demonstrated significant improvements over baseline models and achieved state-of-the-art (SOTA) performance on several tasks, surpassing GPT-4 vision models~\cite{gpt4v, gpt4o}. These results underscore the model's substantial advantages in chemical image recognition and image-text MMCR tasks.

Our key contributions are as follows:

\begin{enumerate}
    \item \textbf{A multimodal LLM dedicated to chemistry.} We propose ChemVLM, an open-source MLLM for the chemistry area. In detail, We present and implement a Vision Transformer-Multi-Layer Perceptron-Large Language Model (ViT-MLP-LLM) architecture specifically tailored for chemical domain visual-language processing.
    \item \textbf{A comprehensive dataset suite.} We develop three new datasets (ChemOCR, MMCR-Bench, and MMChemBench) with diverse tasks to evaluate the capability of visual-language models in the chemical domain.
    \item \textbf{A variety of reliable evaluation.} We assess the performance of different models on various chemical tasks, including benchmarks collected by ourselves and the open-sourced ones. Results show that our model exhibits desirable performance in multimodal chemistry understanding and reasoning.
\end{enumerate}

We hope ChemVLM can advance chemical research by bridging the gap between multimodal data interpretation and domain-specific knowledge and inspire future works in this field. 

\begin{figure}[htbp]
    \centering
    \includegraphics[width=\linewidth]{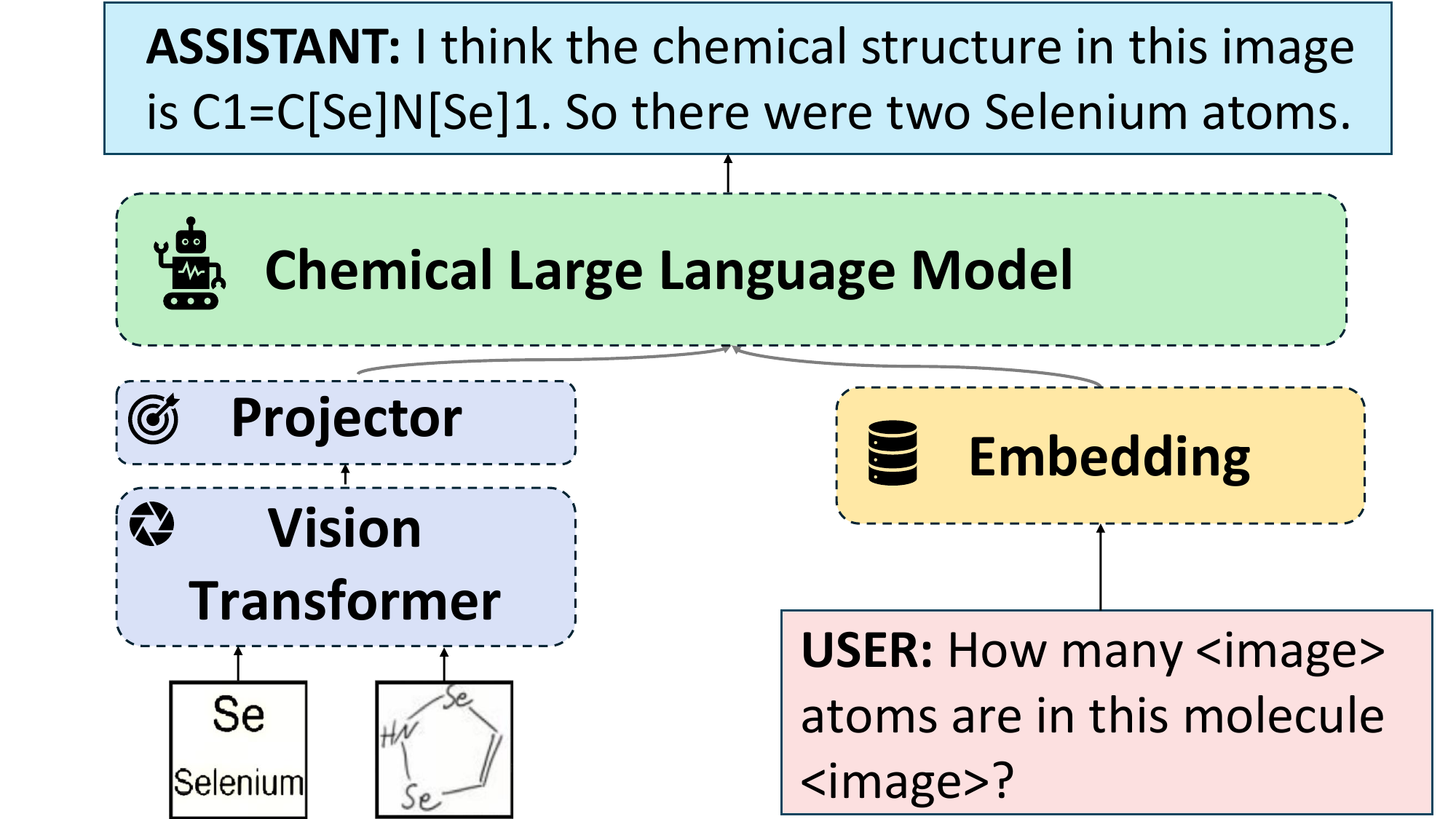}
    \caption{Overall architecture of ChemVLM. ChemVLM combines the advantage of an advanced vision transformer and a large language enriched with chemical knowledge, ensuring the strong ability of multimodal chemical knowledge understanding and reasoning.}
    \label{fig:architecture}
\end{figure}

\section{Related Work}
\textbf{Overview of Multimodal Large Language Models. }
In addition to textual content, visual information encompasses a wealth of data, and integrating these modalities through large language models (MLLMs) has demonstrated superior comprehension, reasoning, and generation capabilities compared to traditional models. Notable MLLMs include GPT-4V~\cite{gpt4v}, which extends GPT-4 with a visual module, and the Gemini series~\cite{gemini,gemini1.5}, which processes audio, video, and text inputs with enhanced long-context understanding. Proprietary models like QwenVL-Plus/Max~\cite{qwenvl}, MM1~\cite{mm1}, and Claude-3V~\cite{Claude3v} also exhibit exceptional multimodal performance. Open-source MLLMs, such as the LLaVA series~\cite{llava,llavanext}, VisionLLM~\cite{visionllm}, CogVLM~\cite{cogvlm}, LLaMA-Adapter V2~\cite{llamaadapterv2}, and ShareGPT4V~\cite{sharegpt4v}, continue to evolve but still lag behind their proprietary counterparts in performance.

\textbf{Overview of Chemical OCR and MMCR.}
Various ways have been attempted by prior works to extract visual information from the chemical images. Works on diagram parsing focused on the segmentation of molecular images~\cite{ImageSeg1,ImageSeg2} and the recognition of their chemical structures~\cite{ImageRecog1,ImageRecog2,ImageRecog3}. Some researchers want to understand the relationships between the molecules, i.e., reaction schemes. Wilary and Cole proposed ReactionDataExtractor~\cite{ReactionDataExtractor} to extract reaction schemes from the diagrams. A method called Chemgrapher~\cite{Chemgrapher} suggests dealing with the problem in a modular fashion by using a segmentation algorithm to segment the images containing chemical graphs to detect atom locations, bonds, and charges.  Only a few try to predict the properties of molecules. Chithrananda et al. make one of the first attempts to systematically evaluate transformers on Multimodal Molecule Understanding tasks such as molecular property prediction~\cite{ChemBERTa}. We provide an MLLM method to help solve the problem of chemical OCR and can work as a chemical properties answer assistant.

\section{Architecture of ChemVLM}

As shown in Figure \ref{fig:architecture}, ChemVLM follows the architecture from LLaVA~\cite{llava,llavanext} in the fashion of "ViT-MLP-LLM". This framework integrates a Vision Transformer (ViT) followed by Multi-Layer Perceptron (MLP) components, seamlessly connected to a Large Language Model (LLM). We adopt InternViT-6B~\cite{chen2024internvl} as the vision encoder and ChemLLM-20B~\cite{zhang2024chemllm} as the language model, as it demonstrates strong chemical capabilities. The projector is an MLP to convert the visual feature into the language embedding space, whose weights are randomly initialized. 
The visual encoder takes images of $448\times448$ resolution as input and extracts high-level features from them. The visual features are then transformed into the same dimension of the language tokens, aligning them with the text space. Simultaneously, the textual input passes through the tokenizer to be converted into a sequence of tokens. The image and text tokens are then concatenated to form a unified token sequence and then fed into the LLM. The LLM synthesizes the multimodal tokens to generate a cohesive and contextually relevant response. This pipeline enhances the model's capability in complex chemical environments. 

\section{Data Composition}

This section details our approach to enhancing our model's recognition and understanding of images of molecules and reactions, and multimodal reasoning tasks in chemistry. We aim to improve the model's capability in real-world multimodal scenarios by leveraging diverse datasets and educational resources. We visualize the data preparation work in Figure~\ref{fig:data}. The data production process is as follows. (1). We collect raw data encompassing various tasks from reliable open-source chemical datasets and databases. (2). We apply diverse image transformation techniques for data augmentation, including style transformations (e.g., handwritten styles, graph paper backgrounds), rotations, blurring, and other enhancement techniques. (3). A variety of templates are carefully designed to cover different task scenarios and improve diversity (please refer to the Appendix for template details). (4). These templates are filled with raw data to create high-quality QA pairs, which will be used to train and evaluate MLLMs. Here we show the data distribution of our train data and benchmarks in Figure~\ref{fig:data-dis}. Detailed explanation can be found in the Appendix.
\begin{figure}
    \centering
    \includegraphics[width=1\linewidth]{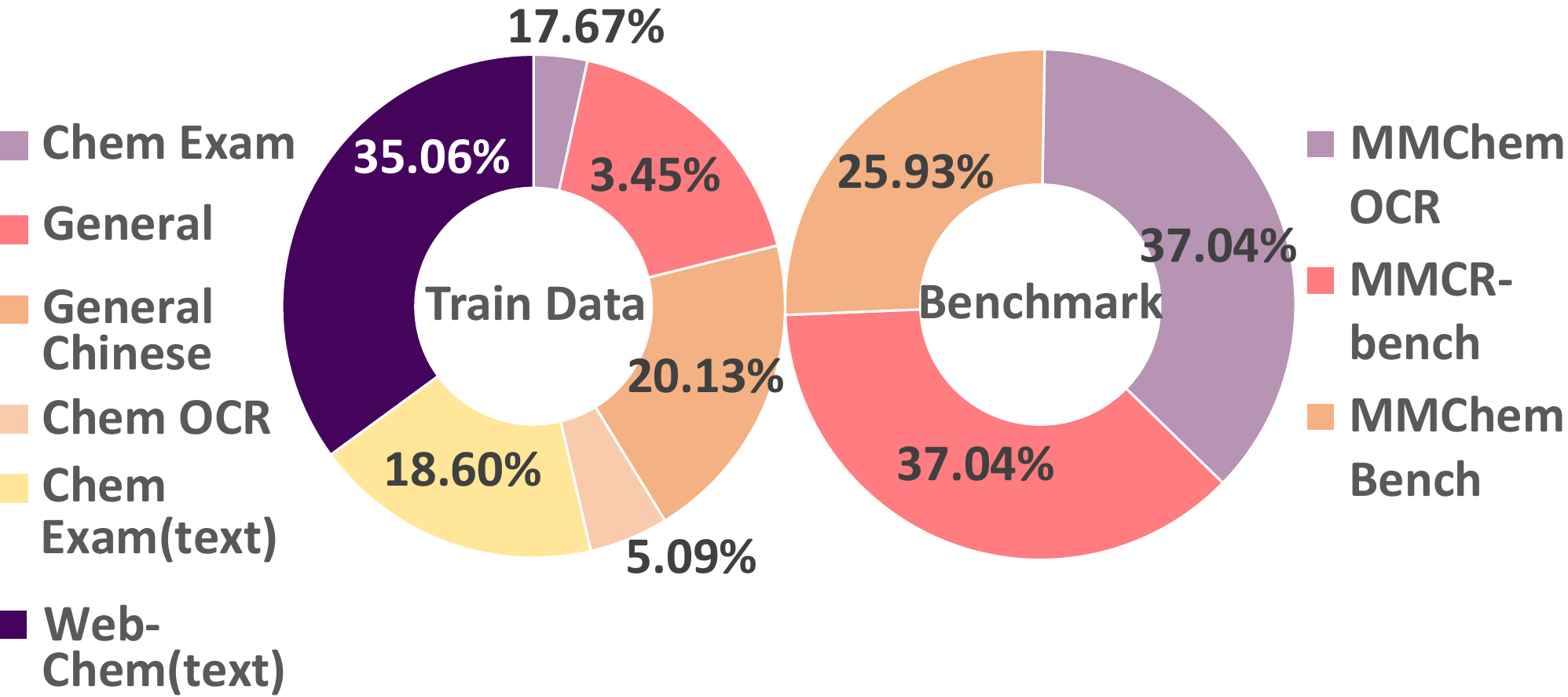}
    
    \caption{Data distribution of our train data and benchmarks.}
    \label{fig:data-dis}
\end{figure}

\subsection{Molecule Subset}

To improve the model's ability on Multimodal Molecule Understanding and Recognition, we collect images from different sources to cover a broad range of styles. The sources are demonstrated as follows. 

\begin{itemize}

    \item \textbf{Hand-drawn style.} We utilize the DECIMER HDM~\cite{brinkhaus_2023_7617107} dataset that contains over 7,000 hand-drawn molecular structure images. This dataset includes drawings created with different inks and types of pens.
    \item \textbf{Scanned and photographed style.} This subset encompasses images with potential artistic enhancements and photographic distortions. We used the dataset compiled by Molscribe Project~\cite{ImageRecog1}, including molecular images sourced from published literature.
    \item \textbf{Electronic document style.} This subset is characterized by images generated by chemical drawing tools, including ChemDraw~\cite{brown2014chemdraw}, RDKit~\cite{landrum2013rdkit} and  Indigo\footnote{A universal cheminformatics toolkit, utilities and database search tools can be found in https://github.com/epam/Indigo}. The images are created with varying line styles and color presets.

\end{itemize}

\subsection{Reaction Subset}

To enhance our model's visual understanding of a broad spectrum of chemical reaction schemes and synthesis processes, we have gathered a diverse collection of images, ranging from simple reaction images to complex synthesis pathway diagrams. The sources are demonstrated as follows. 

\begin{itemize}
    \item \textbf{Simple chemical reaction images.} This subset contains two types of images. The first consists of inorganic chemical reaction equations, primarily sampled from the PEACE~\cite{zhang-etal-2024-peace} dataset, which includes reaction equation image fragments extracted from the literature. The second focuses on organic chemical reactions, which are mainly based on reaction scheme images collected from organic chemistry literature~\cite{Qian2023RxnScribeAS}. Additionally, we synthesize a batch of reaction scheme images using RDKit~\cite{landrum2013rdkit} with data from the USPTO-50K~\cite{Maziarz2024} dataset.
    \item \textbf{Synthesis pathway diagrams.} This subset also includes two types of images. The first features simpler synthesis routes with fewer than five steps gathered from real literature~\cite{Qian2023RxnScribeAS}. The second involves more complex diagrams called Total Synthesis, which aims to create complex large molecules, typically natural products found in biological organisms, from commercially available small molecules artificially. The data are collected from The Organic Chemistry Data~\cite{Myers_2024} and SynArchive website\footnote{https://synarchive.com/}.
\end{itemize}
\begin{figure*}[htbp]
    \centering
    \includegraphics[width=1\linewidth]{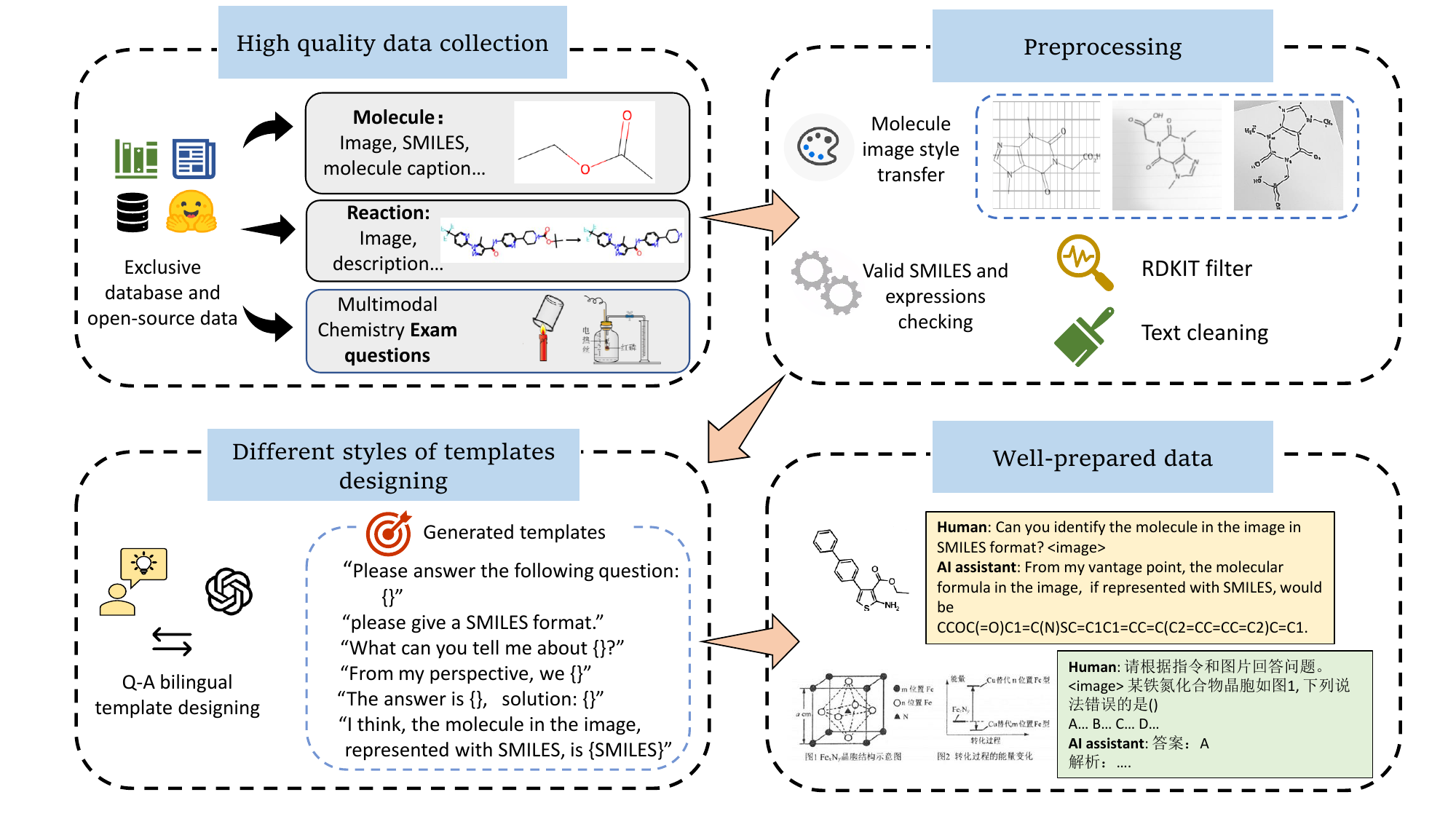}
    \caption{Overview of our data composition work. This multi-step process ensures our model's good performance and a comprehensive evaluation. }
    \label{fig:data}
\end{figure*}

\subsection{Multimodal Chemical Reasoning (MMCR) Tasks}
Existing multimodal chemical models~\cite{ImageRecog1,ImageRecog2,ImageRecog3,Qian2023RxnScribeAS,zhang-etal-2024-peace} primarily focused on modality conversion tasks, such as OCR and molecule captioning, and have achieved a remarkable progress. However, with large language models increasingly being used as tools for knowledge reasoning and scientific discovery~\cite{mirza2024large}, the need for a deeper, more integrated understanding of multimodal knowledge has become critical. Some multimodal question-answering datasets~\cite{mirza2024large,luo2023biomedgptopenmultimodalgenerative} have been involved in the field of chemistry, however, these efforts often focus solely on evaluating the model's ability to recognize and statistically analyze local structures or sub-patterns within non-text modalities, such as identifying specific functional groups within molecular structures. 
For tasks like solving chemistry problems in high school chemistry exams, the model needs to possess strong logical reasoning skills, be able to capture critical information from the questions, and combine it with extensive domain-specific knowledge to provide accurate and reasonable answers.

Given the importance of 
the problem above, we utilize examination data within the chemistry discipline to build a Q\&A dataset. Inspired by C-MHChem~\cite{zhang2024chemllm}, we carefully curate and deduplicate a dataset containing 200,000 high-quality multimodal chemistry questions from the OpenDataLab platform~\cite{he2024opendatalabempoweringgeneralartificial}, covering Chinese education from secondary school to graduate levels\footnote{The chemistry QA pairs from OpenDataLab are originally in Chinese, and we translate 10,704 of them into English using the Bing translator from pypi/translators.}. 
It includes various problem types such as multiple choice, fill-in-the-blank, and short answer. The questions are designed to assess diverse skills of test-takers, including error correction, knowledge-based Q\&A, complex reasoning, and experiment protocol design. This dataset aims to equip the model with a deeper understanding of chemistry, enabling it to tackle complex multimodal challenges in real-world scenarios.

\subsection{Data Quality Enhancement}
To ensure higher data quality and improve the effectiveness of training, we implement a series of data quality improvement techniques, as depicted in Figure \ref{fig:data}. Considering the importance of diverse training data for learning meaningful image-text interactions, we apply RanDepict~\cite{brinkhaus2022randepict} to incorporate variation in image styles. This includes style transformations (e.g., handwriting style, grid paper backgrounds), rotations, blurring, and other data augmentation techniques for molecular images. Additionally, we perform validation checks on the SMILES fields in the text information and remove any irrelevant symbols to maintain accuracy. Furthermore, we diversify the question-answer prompt templates by interacting with GPT-4~\cite{gpt4} to generate prompts in different linguistic styles, both in Chinese and English. These prompts are then used to construct the final question-answer pairs for supervised finetuning.

\section{Training}

\subsection{Training Strategy}

ChemVLM’s training follows a two-stage paradigm inspired by InternVL~\cite{chen2024fargpt4vclosinggap}. The first stage focuses on aligning the image and text modalities, while the second stage involves supervised finetuning. During training, we segment images into pixel tiles with a maximum of 12, depending on the image's aspect ratio and resolution. A context length of 2048 tokens is utilized to accommodate detailed responses, and the response formatting follows the prompt structure used in LLaVA 1.5~\cite{llavanext}.

\begin{table}[b]
  \small 
  \setlength{\tabcolsep}{4pt} 
  \renewcommand{\arraystretch}{1.1} 
  \centering
  \begin{tabular*}{1\linewidth}{@{}l|c|c@{}}
    \toprule[1.2pt]
    \hline
    \textbf{Model\&Method}  & \textbf{Avg Sim. (\%↑)} & \textbf{Tani@1.0 (\%↑)}  \\
    \hline
    Decimer   & 85.0 &  77.3\\
    Molscribe  & 92.0 & 89.1  \\
    \hline
    Qwen-VL-Chat   & 5.0  & 0.0  \\ 
    LLaVA-v1.5-13B   & 1.0 & 0.0 \\
    InternVL-v1.5    & 9.0 & 0.0  \\
    Yi-VL-Plus   & 5.0 & 0.0 \\
    GPT-4V  &  15.0 & 2.1  \\
    ChemVLM-26B \textbf{(ours)}  & \textbf{71.0} & \textbf{42.9}  \\
    \hline
    \bottomrule[1.2pt]
  \end{tabular*}
  \caption{Results on ChemOCR. 
  Tanimoto similarities are written as Avg Sim. (\%↑), and Tanimoto@1.0 written as Tani@1.0 (\%↑). High is better for both metrics.}
  \label{OCR-res}
\end{table}

\noindent
\textbf{Image-Text Modal Alignment Training.} 
In the first stage, we utilize a diverse multimodal dataset, encompassing both a general-purpose corpus and a specialized chemical corpus, to enhance the alignment of visual and textual representations. Specifically, we freeze the weights of LLM and the base visual encoder, training only the randomly initialized projector and the additional LoRA~\cite{hulora} layers in the visual encoder to imbue it with an understanding of chemical space. Finetuning with LoRA effectively reduces the number of trainable parameters and computational costs, while simultaneously lowering the risks of model overfitting and catastrophic forgetting.

\noindent
\textbf{Supervised FineTuning Training.} 
In the second stage, we finetune the model using a combination of multimodal data and pure text data. The training dataset includes both specialized data from the field of chemistry and general-purpose corpora. 
In this process, since ViT and LLM account for a significant proportion of the parameters, LoRA is applied to finetune the LLM and ViT components to reduce training costs. Meanwhile, the projector, which has a much smaller parameter footprint, is fully trained using all parameters. The parameters used in this stage are derived from those merged and processed in the first stage.

\subsection{Training Details}
For each stage, the model is trained on 16 NVIDIA A100$\times$80G GPUs for one epoch. The batch size is set to 4 and gradients are accumulated over 4 iterations. 
We use AdamW~\cite{loshchilov2017adamw} as the optimizer and Deepspeed bfloat16 (bf16)~\cite{rasley2020deepspeed} precision for efficient training. To handle distributed training, we consistently apply the Deepspeed ZeRO-3~\cite{rajbhandari2020zeromemoryoptimizationstraining} strategy. We adopt the chat template from InternLM2~\cite{internlm} as the dialogue schema for the LLMs. Please refer to the supplementary material for more detailed information. 

\section{Experiments}
In this section, we conduct extensive experiments to assess the chemical capabilities of ChemVLM and other competing models on benchmarks of various tasks. 



\begin{table}[t]
  \small 
  \setlength{\tabcolsep}{1mm} 
  \renewcommand{\arraystretch}{1.3} 
  \centering

  \begin{tabular*}{1\linewidth}{@{}@{\extracolsep{\fill}}l|c|c|c}
    \toprule[1.2pt]
    \hline
    \textbf{Model} & \textbf{ScienceQA}  &  \textbf{CMMU} &\textbf{MMCR- } \\

    \textbf{\&Method} & \textbf{Chem (\%↑)}  &  \textbf{Chem (\%↑)} &\textbf{Bench (\%↑)} \\
    
    \hline
    Qwen-VL-Chat  & 65.3 & 22.1 & 28.3  \\
    LLaVA-v1.5-13B  & 43.6 & 18.6 & 23.1\\
    InternVL-v1.5  & 74.0 & 26.9 & 36.5 \\
    Yi-VL-Plus  & 68.1 & 29.1 & 35.3\\
    GPT-4V   &  \textbf{71.9}  & 24.2 & 40.1\\
    ChemVLM-26B \textbf{(ours)}  & 71.2 &  \textbf{31.6} &\textbf{41.7}\\
    \hline
    \bottomrule[1.2pt]
  \end{tabular*}
  \caption{\textbf{Results on Chemisty part of CMMU, ScienceQA, and MMCR-Bench.} We compare our model's performance with several MLLMs and report the accuracy. The evaluation is conducted in a zero-shot manner. }
  \label{QA-bench-res}
\end{table}

\subsection{Evaluation Settings}
To assess the multimodal capabilities of models on chemical tasks, we evaluate their performance on both publicly available and self-curated benchmarks. 
Our evaluation focuses on the three critical chemical tasks, including Chemical Optical Character Recognition (Chemical OCR), Multimodal Chemical Reasoning (MMCR), and Multimodal Molecule Understanding. 

We utilize the chemistry part of open-source multimodal benchmarks, CMMU~\cite{he2024cmmu} and ScienceQA~\cite{lu2022learn} for MMCR evaluation. 
To evaluate the model's capability of Multimodal Molecule Understanding, we construct a multimodal chemical benchmark named \textbf{MMChemBench}, derived from ChemBench~\cite{zhang2024chemllm} with two image-text tasks: molecule caption and molecular property prediction. Meanwhile, we employ other scientific disciplines within the CMMU~\cite{he2024cmmu} framework to assess ChemVLM's generalization capabilities in different areas. We also adopt Scibench~\cite{wang2024scibench}, a college-level chemistry test set consisting of purely textual data, to evaluate the model's problem-solving abilities in more complex tasks.

Moreover, we build two datasets for a more comprehensive benchmarking called \textbf{ChemOCR} and \textbf{MMCR-Bench}. Specifically, ChemOCR includes 1000 chemical OCR image-text pairs collected from an open-source chemical database~\cite{ImageRecog1}. And MMCR-Bench includes 1000 high-quality multimodal chemistry exam questions from the Chinese college entrance examination. 

\subsection{Results on Chemical OCR Task}\label{ocr-sec}
In this task, the models are expected to generate a corresponding SMILES string for each molecular image. 
We compare our model with previous end-to-end models dedicated to this task, including Decimer~\cite{rajan2021decimer} and Molscribe~\cite{ImageRecog1}. These models are limited to generating SMILES and lack natural language processing capabilities. Therefore, the comparison is also conducted on various multimodal LLMs~(MLLMs) including Qwen-VL-Chat~\cite{bai2023qwen}, LLaVA-v1.5-13B~\cite{llava}, InternVL-v1.5~\cite{chen2024far}, Yi-VL-plus~\cite{young2024yi}, and GPT-4V~\cite{openai2024gpt4o}. We employ 1000 image-text pairs in ChemOCR to evaluate the model’s overall chemical OCR capabilities. We compute the Tanimoto similarity between the generated molecules and ground truth molecules with RDKIT~\cite{landrum2013rdkit}, and report the average Tanimoto similarity and Tanimoto hit 1.0 (tanimoto@1.0), which measures the percentage of structures with 1.0 similarity. 


The evaluation results are shown in Table~\ref{OCR-res}. Our model exhibits strong performance on this task, outperforming all competing MLLM models. While ChemVLM's performance is behind specialized models like Decimer and MolScribe, it offers greater versatility in handling a wider range of tasks, not limited to Chemical OCR. Notably, a 59\% increase in average similarity and a 40.8 increase in tanimoto@1.0 can be found when compared with GPT-4V. 



\begin{figure}[!t]
    \centering 
    \includegraphics[width=\linewidth]{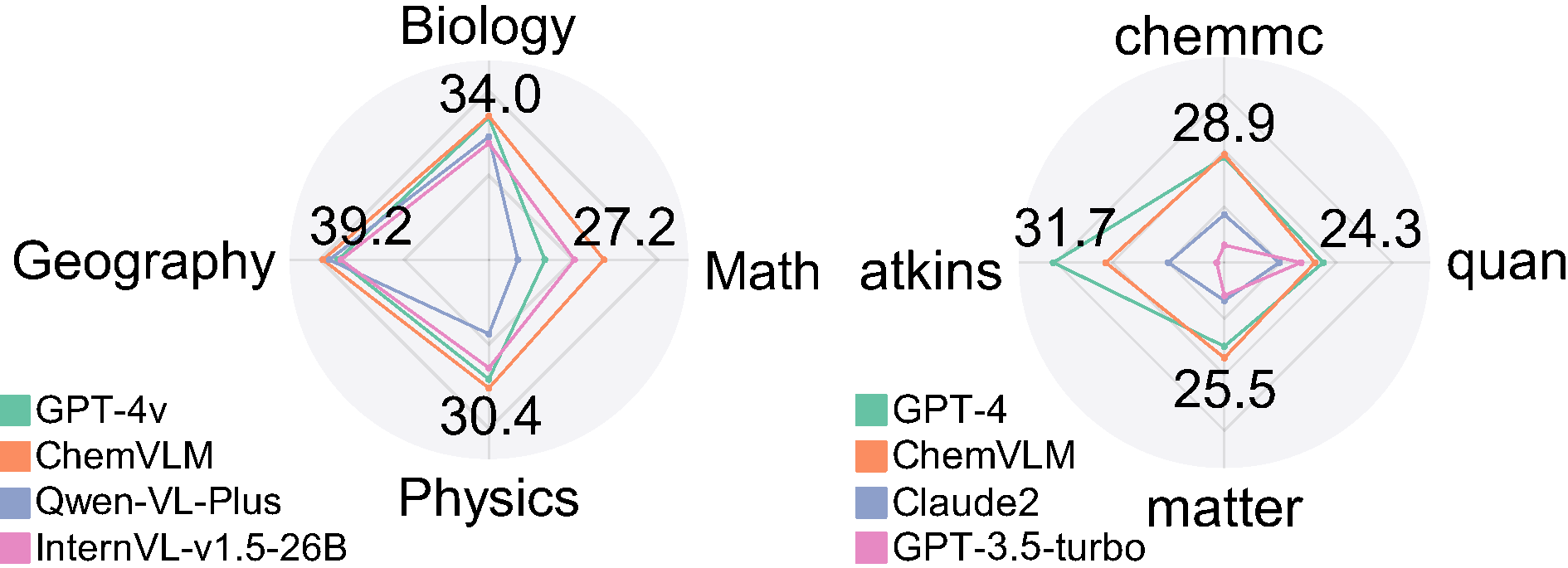}
    \caption{In the left figure, we compare ChemVLM with three other MLLMs on other subjects aside from chemistry on CMMU. In the right figure, we show results on the subsets related to chemistry on Scibench. The numbers represent the performance of ChemVLM.}
    \label{fig:cmmu_other_subjects}
\end{figure}

\begin{figure*}
    \centering
    \begin{minipage}[t]{0.495\textwidth}
    \centering
    \includegraphics[width=\linewidth]{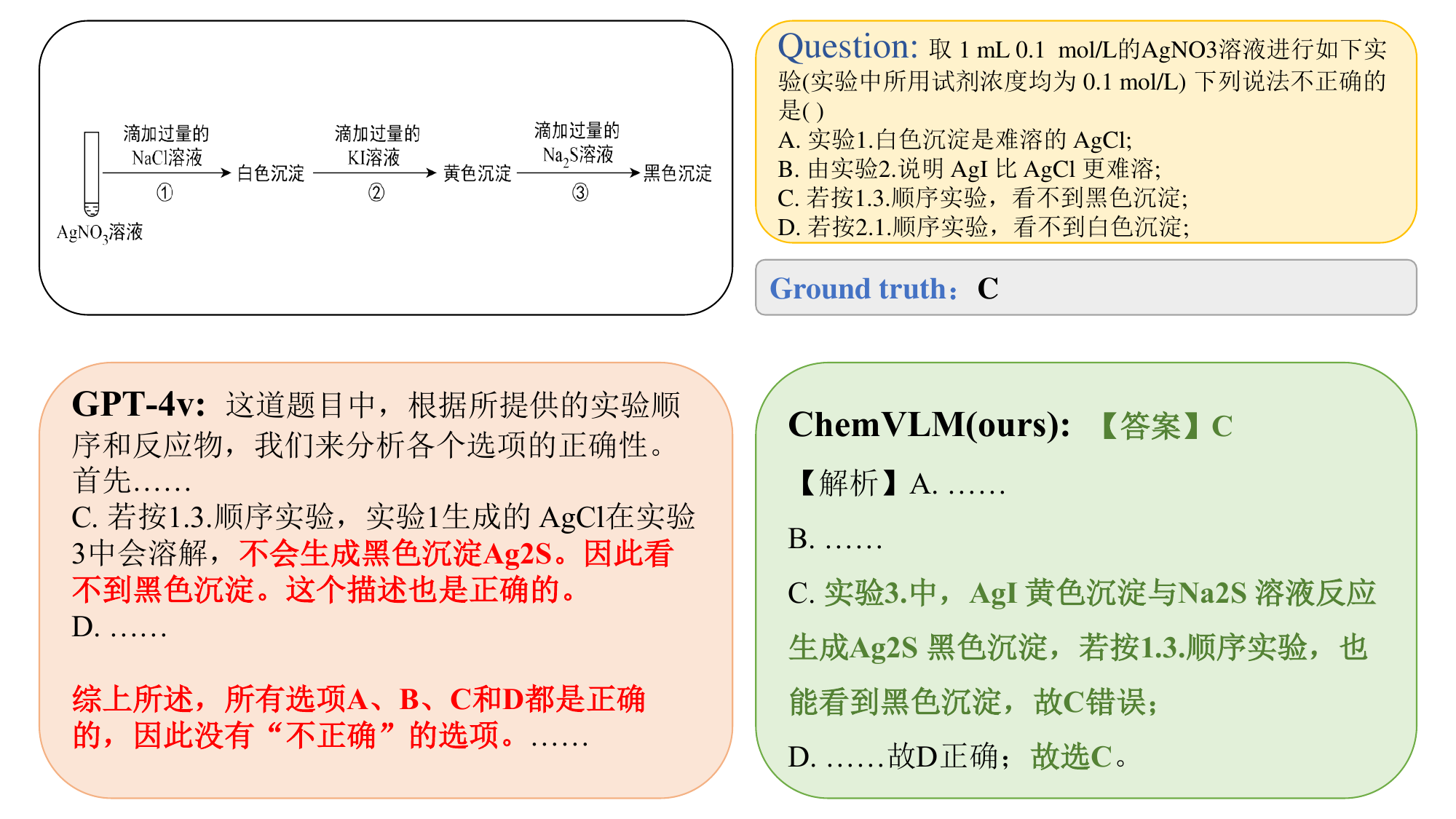}
    \label{fig:1}
    \end{minipage}
    \hfill
    \begin{minipage}[t]{0.495\textwidth}
    \centering
    \includegraphics[width=\linewidth]{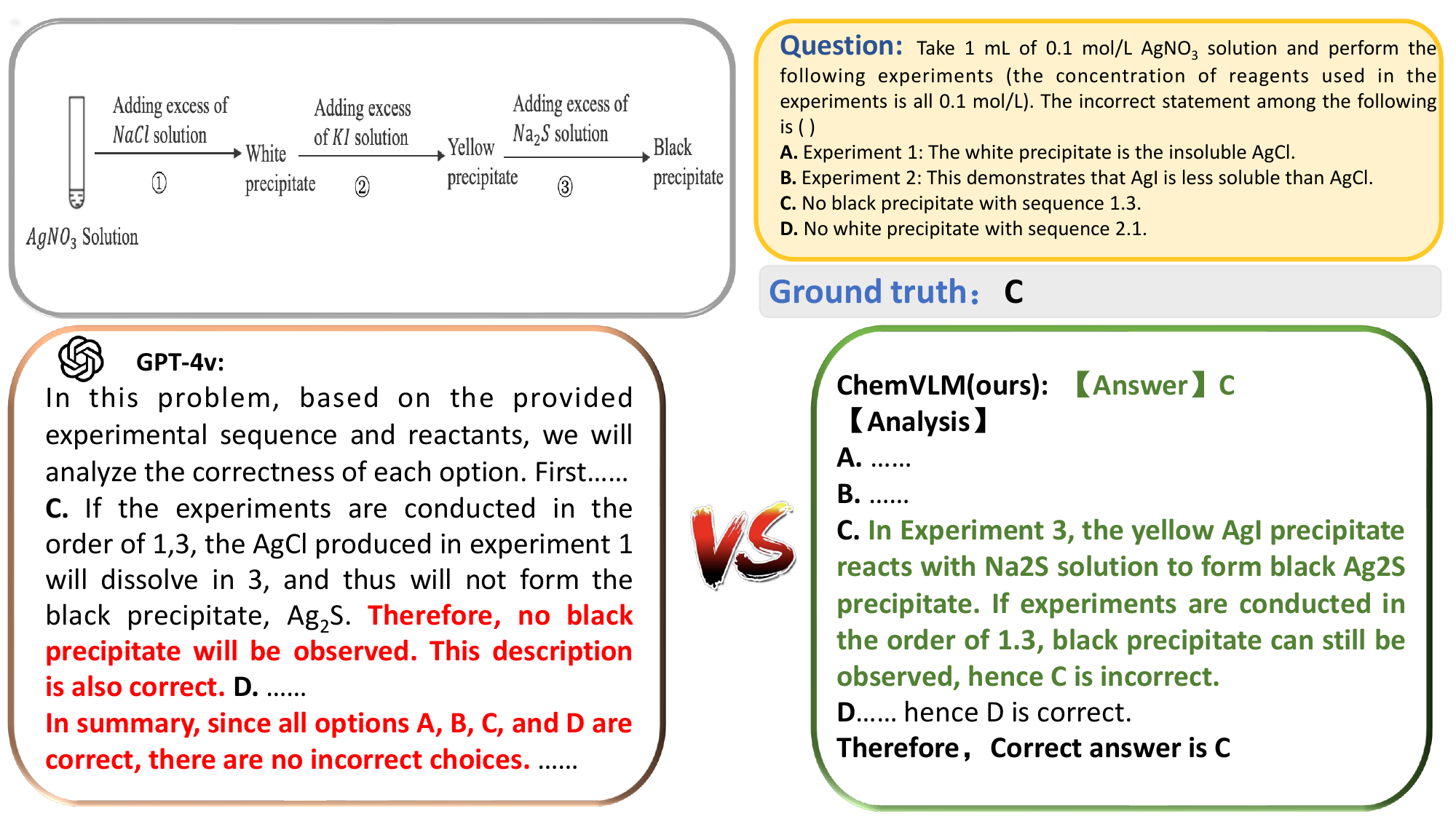}
    \label{fig:2}
    \end{minipage}
    \caption{A qualitative comparison of answers on MMCR-Bench between GPT-4V and our ChemVLM.  Mistakes within the answers are highlighted in red, whereas detailed and accurate parts are emphasized in green. Since this is a Chinese exam question, we prepare the original Chinese text and the English translation of it. \textbf{This shows the strong MMCR capability of ChemVLM.}}
    \label{fig:visual1}
\end{figure*}

\subsection{Results on MMCR Tasks}
In addition to the OCR-related tasks, we test our model on question-answer tasks compiled from exams. The task involves data from three different sources. We utilize the chemical sections of the CMMU and ScienceQA datasets, which are collected from exams across various grade levels. Additionally, we include 1000 multimodal chemistry exam questions from our custom test set, MMCR-Bench. The total score is calculated based on the following criteria. 

\textit{An answer gets one point when it matches all the right choices of a \textbf{multiple-choice problem}, else it gets zero point. }

\textit{For a \textbf{fill-in-the-blanks problem}, an answer should make all the blanks right, and then get one point. A wrong blank means zero points for the whole problem. We complete this process by prompting the Qwen-max API~\cite{yang2024qwen2}. Please refer to the supplement for more details.}

\textit{At last, the total score is calculated by dividing the number of questions into points a model gets. }

Results on the MMCR tasks are summarized in Table \ref{QA-bench-res}. Our model achieves state-of-the-art (SOTA) for the chemical part on the CMMU benchmark. 
Additionally, ChemVLM demonstrates strong performance in the chemistry section of ScienceQA, closely matching that of GPT-4V. Upon analyzing the ScienceQA dataset, we find that many of the questions rely on common sense, making them easier for MLLMs to solve.  When it comes to MMCR-bench, ChemVLM shows more promising performance, which surpasses GPT-4V by 1.7\% and achieves state-of-the-art performance. 

\begin{table}[htbp]
  \small 
  \setlength{\tabcolsep}{1mm} 
  \renewcommand{\arraystretch}{1.5} 
  \centering

  \begin{tabular*}{1\linewidth}{@{}@{\extracolsep{\fill}}l|c|c}
    \toprule[1.2pt]
    \hline
    \textbf{Model}  & \textbf{molecule} & \textbf{property}   \\
    \textbf{\&Method} & \textbf{caption(\%)} & \textbf{prediction(\%)} \\
    \hline
    ChemLLM\textbf{*}  & 92.6 & 72.2 \\
    GPT-4\textbf{*}   & 96.3 & 68.4 \\
    \hline
    Qwen-VL-Chat  & 86.6 &  56.1 \\  
    LLaVA-v1.5-13B  & 75.9 & 19.6 \\
    InternVL-v1.5  & 78.7 &  32.2 \\
    Yi-VL-Plus  & 89.3 & 27.4 \\
    GPT-4V  &  95.2 & 38.6 \\
    ChemVLM-26B \textbf{(ours)} & \textbf{98.2} & \textbf{80.9} \\
    \hline
    \bottomrule[1.2pt]
  \end{tabular*}
  \caption{\textbf{Results on MMChemBench.} We report the total score using the method of calculating a score for choice problems. Notably, we report text-only LLMs'(*) performance on MMChemBench's corresponding data source: ChemBench.}
  \label{QA-multimodal-res}
\end{table}

\subsection{Results on Multimodal Molecule Understanding Tasks}
We present MMChemBench, an extension of ChemBench, designed to evaluate our models' performance in Multimodal Molecule Understanding tasks which include molecule caption and molecule property prediction. 
We also compare ChemVLM with ChemLLM and GPT-4 to assess the impact of incorporating visual modality into the model. The results are presented in Table~\ref{QA-multimodal-res}, which demonstrate that ChemVLM achieves state-of-the-art performance in both tasks. A significant improvement is observed in the property prediction task compared to other MLLMs. The findings highlight that incorporating visual information greatly enhances the model's ability to understand domain-specific knowledge, such as molecular structures.

\begin{figure}[h]
    \centering \includegraphics[width=0.9\linewidth]{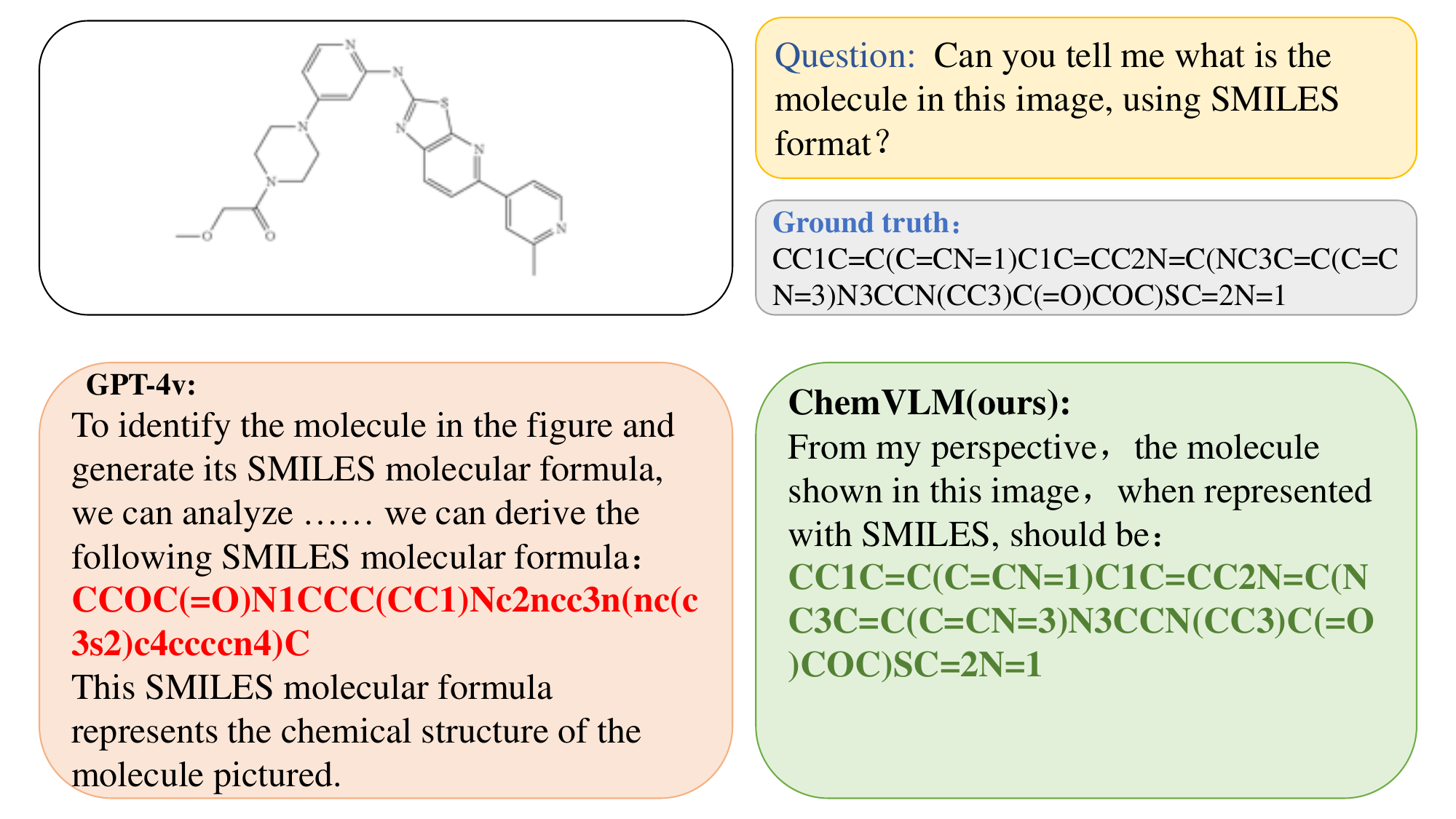}
    \caption{A qualitative comparison of answers on ChemOCR on GPT-4V and our ChemVLM. Red and green parts are the same as Figure~\ref{fig:visual1}.}
    \label{fig:visual2}
\end{figure}


\subsection{Study on Multidisciplinary Competence}
Our proposed ChemVLM is primarily trained on a proprietary chemistry dataset. To evaluate the model's generalization capabilities, we tested it on test sets from other scientific fields within the CMMU framework. As shown in Figure~\ref{fig:cmmu_other_subjects}, the experimental results indicate that ChemVLM performs exceptionally well in other disciplines, surpassing renowned multimodal LLMs and InternVL-v1.5-26B. This suggests that large-scale training on high-quality problems enables our model to exhibit remarkable capabilities across different scientific domains. 

\subsection{Study on More Difficult Questions}
To test the model's reasoning capability on complex chemistry problems, we evaluated various models using the Scibench~\cite{wang2024scibench} dataset. Scibench is a \textbf{text-only} benchmark designed for university-level science questions extracted from exercise books, including topics like quantum chemistry and physical chemistry. As shown in Figure~\ref{fig:cmmu_other_subjects}, our model outperformed all models mentioned in their paper, except for the GPT-4 series. Notably, in the "Chemmic" and "Matter" subtasks, our model achieved better results than GPT-4. This highlights ChemVLM's reasoning ability on purely textual, more difficult questions, despite being designed to address multimodal chemistry problems.

\subsection{Visualization of Answers From Different MLLMs and Our ChemVLM}

In this part, we present qualitative results to compare the outputs of our model with those from a strong proprietary MLLM, GPT-4V, as shown in Figure \ref{fig:visual1} and Figure \ref{fig:visual2}. The results indicate that our model exhibits a better understanding of the questions and provides more accurate answers for chemical OCR and examination questions. This highlights ChemVLM's proficiency in leveraging chemical knowledge to address multimodal problems.

\section{Conclusion and Future Work}


In this study, we introduce ChemVLM, an open-source, multimodal model tailored for applications in chemistry. Comprehensive evaluations indicate that ChemVLM exceeds most existing methods, advancing the AI and chemistry fields.

Currently, ChemVLM primarily incorporates image modalities and lacks processing capabilities for molecular graph and time-series data, limiting its effectiveness in capturing complex chemical phenomena. Additionally, the high computational costs may hinder its wider adoption.

In the future, we aim to enhance ChemVLM by adding new modalities, such as graph and time-series data. We also plan to explore more efficient training methods and develop a range of models with varying parameter sizes to suit diverse scenarios, thereby advancing LLM's role in chemistry research.

\section*{Acknowledgements}
This work is supported by Shanghai Artificial Intelligence Laboratory.

\bibliography{aaai25.bib}

\clearpage

\pagestyle{empty}

\appendix

\section{Technical Appendix}
Some details of our framework are listed here.

\section{Training settings}
Here we list out the training settings of our two-stage training, for better reproduction. 

\begin{table}[h]
\renewcommand{\arraystretch}{}
    \centering
    \scriptsize
    \setlength{\tabcolsep}{1mm}
    
    \caption{Training settings of ChemVLM's stage 1 and stage 2.} 

    \begin{tabular}{l|cc}
        \hline
        \toprule[1.3pt]
        \hline
        
        \textbf{Settings} & \textbf{Stage 1} & \textbf{Stage 2} \\
        \hline
        \rowcolor{gray!20}
        freeze llm & True & True \\
        llm lora rank & 0 & 16 \\
        \rowcolor{gray!20}
        freeze mlp & False & False \\
        freeze vit & True & True \\
        \rowcolor{gray!20}
        vit lora rank & 32 & 32 \\
        learning rate & 5e-5 & 5e-5 \\
        \rowcolor{gray!20}
        learning rate schedule & cosine decay & cosine decay \\
        optimizer & AdamW & AdamW \\
        \rowcolor{gray!20}
        optimizer hyper-parameters & $\beta_{1}$, $\beta_{2}$ = 0.9, 0.999 & $\beta_{1}$, $\beta_{2}$ = 0.9, 0.999 \\
         weight decay & 1e-5 & 0.00 \\
         \rowcolor{gray!20}
         input size & 448$\times$448 & 448$\times$448 \\
         total batch size & 16$\times$4$\times$4 & 16$\times$4$\times$4 \\
         \rowcolor{gray!20}
         drop rate & 0.0 & 0.0 \\
         numerical precision & DeepSpeed bf16 & DeepSpeed bf16 \\
         \rowcolor{gray!20}
          GPUs for training & 16$\times$A100 (80G) & 16$\times$A100 (80G) \\
          
          \hline
          \bottomrule[1.2pt]
          \hline
    \end{tabular}
\label{tab:train_cfg_stage}
\end{table}

\section{Data generation prompts}
In the main content, we introduce how we generate the chemical data we need. Here we list out the sample prompts, helping understand our work.

Prompts for generating question-answer templates(calling GPT-4):

\textbf{\textit{Can you offer me ten(or more) different expressions of $\{\}$? Remember that you should use as much different words and lingual styles as you can. }}

In $\{\}$ we add one simple expression of templates we need, like "\textbf{\textit{Can you tell me what is this molecule in the image?}}".

Samples of generated templates:

1. (Chemical OCR) \textbf{\textit{Human: Can you explain the composition of the molecule in the image in SMILES style?}}

\textbf{\textit{Assistant: My interpretation is that the molecular formula in this image, according to SMILES format, is $\{$SMILES$\}$.}}
\newline

2. (Molecule caption and property prediction) \textbf{\textit{Human: What can you tell me about this molecule in the image?}}

\textbf{\textit{Assistant: As far as I'm concerned, the molecule in the image, is $\{$Name$\}$. It has $\{$Properties$\}$.}}
\newline

3. (Exam question answering) \textbf{\textit{Human: $\{$A certain \ exam \ question$\}$. Understand the question and the image, and then give your answer.}}

\textbf{\textit{Assistant: I believe, the answer is $\{$Answer$\}$. The analysis to the question is $\{$Detailed 
 \ solutions$\}$.}}
\newline

4. (Reaction analysis) \textbf{\textit{Human: This diagram appears to illustrate a synthesis process; what is the product, and which reagents are utilized?}}

\textbf{\textit{Assistant: This diagram, as I understand it, aims to synthesize $\{$target$\}$, covering the path $\{$path$\}$ and using reagents such as $\{$reagent$\}$.}}

\textbf{...}

Here we only show some samples of our generated templates on certain tasks. Our total data is a mixture of multi-task data in different language styles.

\section{Dataset details}
Here we list out the data details we use for training and evaluation in Table 2 and 3. These data consists of our own data, data downloading from huggingface, science databases, etc. \textbf{Proper citations are added in the References part.} Note that our model use the pretrained parameters from ChemLLM-20B, so it has  abilities of understanding chemical knowledge itself. Additionally, all the data we mention here are multimodal data if a '(*)' is not used. Data with '(*)' are text-only data.

\begin{table}[h]
    \centering
    \caption{Training data details. 'Chem' means this data is chemical data. General means common sense data. Web-Chem means a set of chemical datasets we download online. Datasets printed in \textit{italic type} are open-source datasets.}
    \begin{tabular}{c|c|c}
    \hline
        \toprule[1.3pt]
        \hline
        \textbf{Dataset name} & \textbf{Amount} & \textbf{Dataset Description} \\ \hline
        Chem Exam & 122k & exam questions \\
        \textit{General} & 625k & English general questions \\
        \textit{General-Chinese} & 712k & Chinese general questions \\
        Chem OCR & 180k & Chemical OCR questions \\
        Chem Exam(*) & 658k & text-only exam questions \\
        \textit{Web-Chem(*)} & 1240k & web multi-task data \\ \hline

    \end{tabular}
    \label{tab:training_data}
\end{table}

\begin{table}[h]
    \centering
    \caption{Evaluation data details.  Datasets printed in \textit{italic type} are open-source datasets. }
    \begin{tabular}{c|c|c}
    \hline
        \toprule[1.3pt]
        \hline
        \textbf{Dataset name} & \textbf{Amount} & \textbf{Dataset Description} \\ \hline
        \textit{CMMU} & about 240 & Chinese exam questions \\
        \textit{ScienceQA} & about 200 & English general questions \\
        MMChemOCR & 1000 & SMILES OCR questions \\
        MMCR-bench & 1000 & Chinese exam questions \\
        MMChemBench & 700 & English general questions \\
          \hline

    \end{tabular}
    \label{tab:eval_data}
\end{table}

\section{Evaluation details and prompts}
In this section, we list out the details and prompts we use for evaluating various MLLMs on some task.
\subsection{Evaluating chemical OCR}
In this task, MLLMs need to recognize what the molecule is in a image. And then, they should generate a SMILES name for it. Many MLLMs' answers do not only contain SMILES strings, but analysis for the question. However, the evaluation metrics need SMILES only. Thus, we design prompts to extract the SMILES they generate and use SMILES for score calculation. Note that we design the Chinese(Zh) and English(En) version. Since extracting SMILES is a text-only task, we call the Qwen-max API with this prompt and keep the answers from Qwen-max. The prompts are in figure 1:

\begin{figure}
    \centering
    \includegraphics[width=1\linewidth]{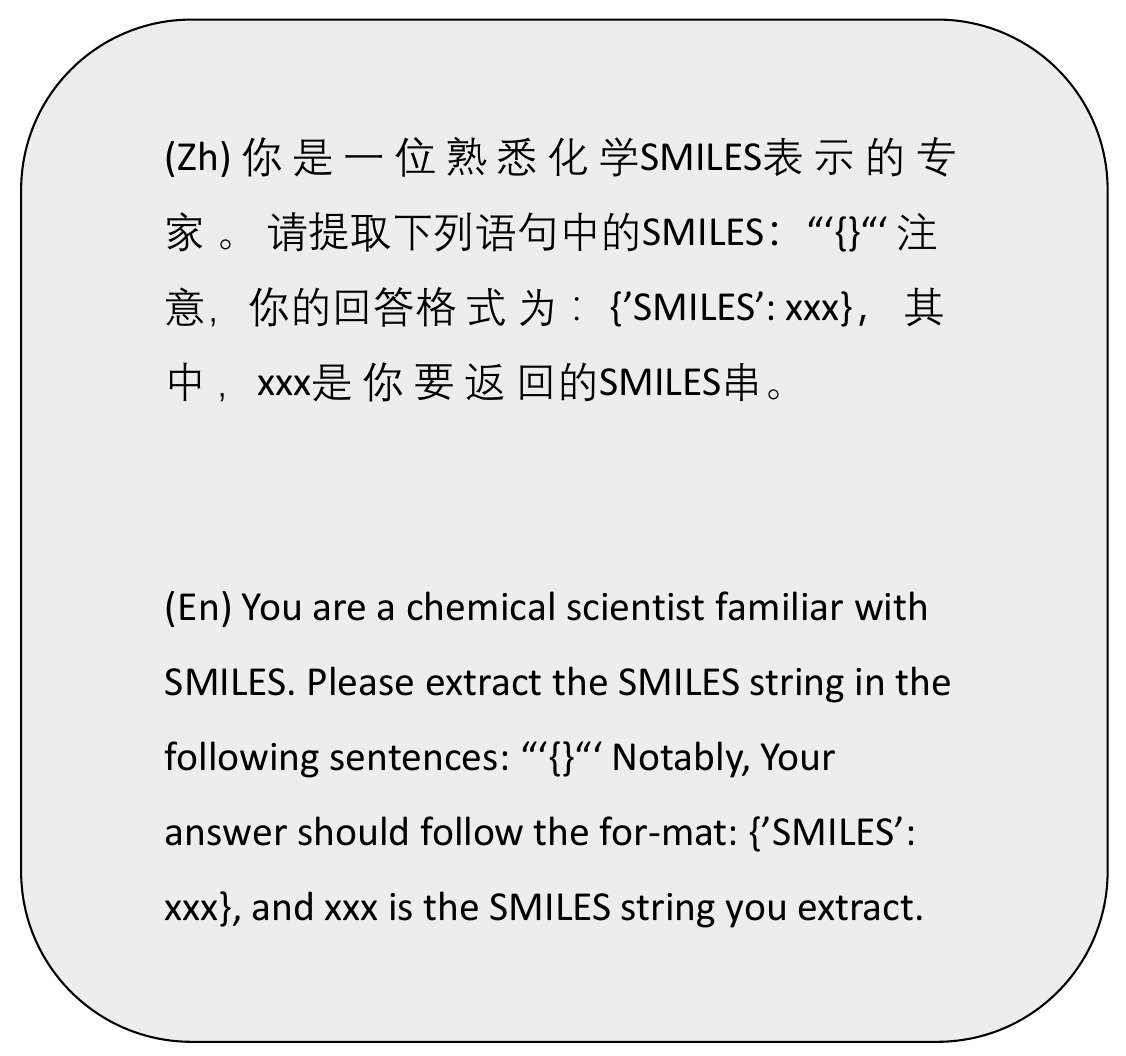}
    \caption{Prompts for extracting SMILES.}
    \label{fig:smiles}
\end{figure}

These prompts all means 'SMILES extraction'. Fill in the $\{\}$ with answer from MLLMs and the prompts are completed. Additionally, we choose Qwen-max for its strong ability for extracting certain texts and convenience during testing various LLMs.
\subsection{Evaluating exam questions}
When we evaluate MMCR-bench, CMMU and ScienceQA, we may face problems. There are quite a few fill-in-the-blanks problems in the benchmarks, so manually counting the number of right blanks is impossible. Thus, we also choose to design special prompts and call Qwen-max API to do these tasks. The prompts are in figure 2:

\begin{figure}
    \centering
    \includegraphics[width=1\linewidth]{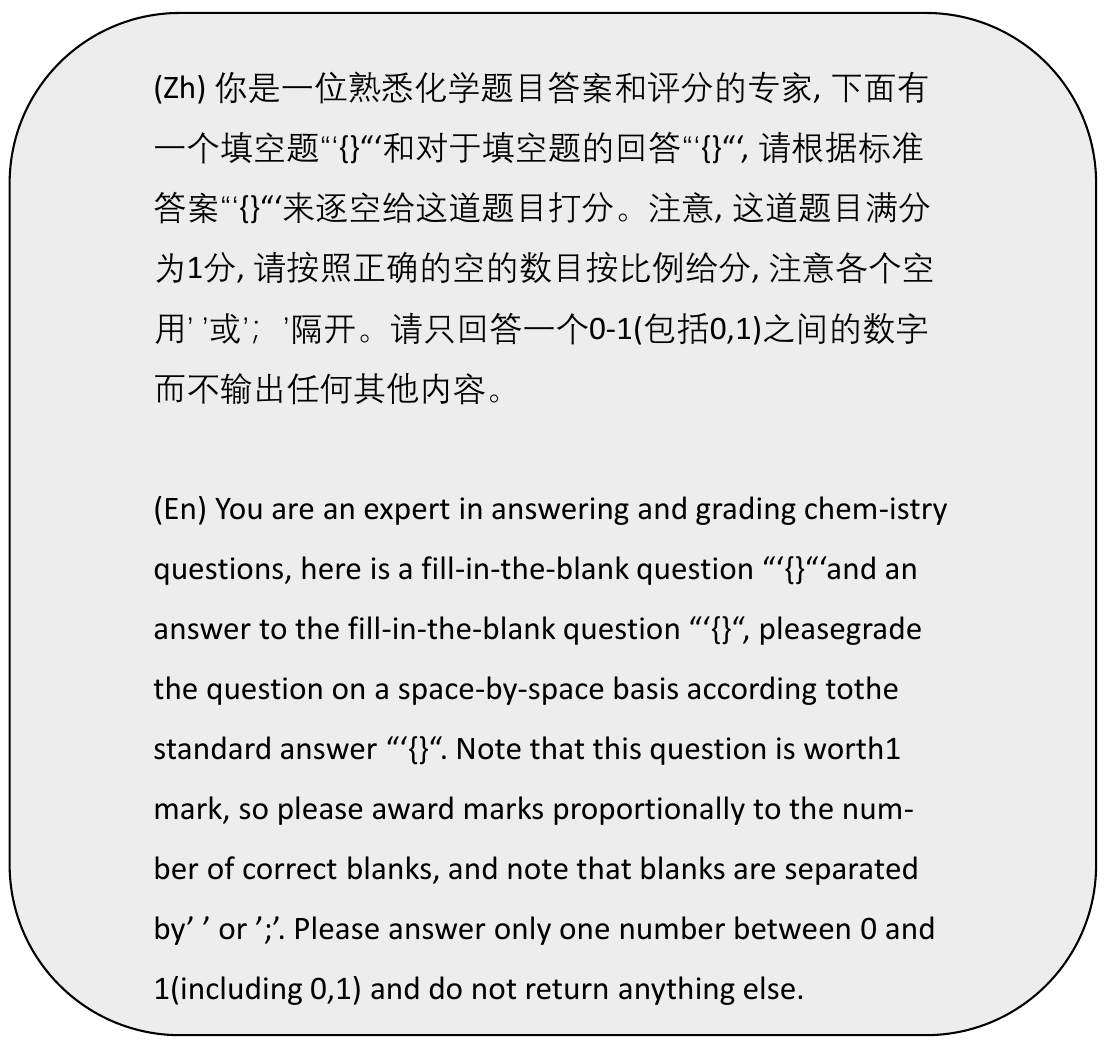}
    \caption{Prompts for fill-in-the-blank questions scoring.}
    \label{fig:score}
\end{figure}

\fontsize{9.5pt}{10.5pt}
\selectfont


\end{document}